\documentclass[12pt]{article}

\usepackage{graphicx}
\DeclareGraphicsExtensions{.pdf,.jpg}
\usepackage{amsmath}
 
\usepackage{tikz}
\usepackage{mathdots}
\usepackage{yhmath}
\usepackage{array}
\usepackage{multirow}
\usepackage{tabularx}
\usepackage{extarrows}
\usepackage{booktabs}
\usetikzlibrary{shapes}
\begin{document}

\title{Learning the P2D Model for Lithium-Ion Batteries with SOH Detection} 

\author{
Maricela Best McKay \thanks{Mathematics, University of British Columbia, maricela@math.ubc.ca} \and Bhushan Gopaluni \thanks{Chemical and Biolgical Engineering, University of British Columbia, bhushan.gopaluni@ubc.ca}
\and Brian Wetton \thanks{Mathematics, University of British Columbia, wetton@math.ubc.ca}
}

\maketitle

\begin{abstract}  
Lithium ion batteries are widely used in many applications. Battery management systems control their optimal use and charging and predict when the battery will cease to deliver the required output on a planned duty or driving cycle. Such systems use a simulation of a mathematical model of battery performance. These models can be electrochemical or data-driven. Electrochemical models for batteries running at high currents are mathematically and computationally complex. In this work, we show that a well-regarded electrochemical model, the Pseudo Two Dimensional (P2D) model, can be replaced by a computationally efficient Convolutional Neural Network (CNN) surrogate model fit to accurately simulated data from a class of random driving cycles. We demonstrate that a CNN is an ideal choice for accurately capturing Lithium ion concentration profiles. Additionally, we show how the neural network model can be adjusted to correspond to battery changes in State of Health (SOH).
\end{abstract}

\section{Introduction}
\label{s:intro}

Rechargeable Lithium Ion Batteries (LIB) are ubiquitous in portable electronics, electric cars, and small scale power storage \cite{LIB_overview,LIB_overview2,LIB_overview3,BatteryU}. Battery Management Systems \cite{BMS,BMS2,BMS3} (BMS) control the charge and discharge of LIB systems, and estimate the State of Charge (SOC) of a system.  Sophisticated BMS also adjust to the battery State of Health (SOH) \cite{BMS_SOH1,BMS_SOH2,SOH3,SOH4,SOH5}, how battery performance changes with time and use. All BMS rely on models of battery performance. These models can be simple ones, based on equivalent circuit approximations of LIB operation \cite{Equiv1}. These are suitable for low current (low C-rate) applications. For the reader unfamiliar with battery units, a ``C'' is the current which will discharge a fully charged battery in an hour. It is well known that LIB operating at high currents need more complicated electrochemical models to accurately predict performance \cite{P2Dreview}. Here, the intercallation rate of Lithium ions in and out of electrode particles can be current limiting. Ionic currents in the electrolyte can also be rate limiting. If both these effects are important, then the well-known Pseudo Two Dimensional (P2D) electrochemical model is needed to predict performance. The P2D model is also known as the Doyle-Fuller-Newman (DFN) model after the researchers that first proposed it \cite{P2D}. The models above can be classified as physical models, with parameters that can be fit from limited experiments and then extrapolated to other operating conditions. Another option for LIB models is modern Deep Learning (DL) \cite{ML1,ML2,ML3}, which relies on large quantities of observed operational data. 

To our knowledge, no BMS currently in operation uses the P2D model for real time prediction and control. This model, which involves several coupled nonlinear Partial Differential Equations (PDE), is just too complex to be solved in real time \cite{bhushan,han2021}. In this article, we show that a (DL) surrogate model can give accurate predictions to a P2D model when trained appropriately on synthetic data from a computational simulation. A key aspect of this work is the use of a CNN model, whose structural features are particularly well suited to capturing Li-ion battery concentration data, enabling the model to achieve high accuracy. This work gives insight into the type of training data necessary to accurately capture high current operation and gives a computationally efficient simulation tool for a P2D model (fit to a particular LIB type) that can be used in a BMS. Because we employ a deep learning framework to replace the P2D model, operational data can be easily added to improve the accuracy of the P2D model fit. This is showcased in \ref{s:SOH} where we demonstrate how to incorporate a SOH parameter into the surrogate modelling framework.  

This work is a continuation of an earlier project described in \cite{maricela0}. In this paper, we extend the earlier work from the simple Single Particle Dynamics (SPD) electrochemical model to the full P2D model and employ a CNN rather than a simple, fully connected Neural Network. We show simulations in our driving cycle framework where the P2D model is significantly more accurate. We also show that a mechanism to identify SOH parameters can be incorporated into the surrogate modelling framework. 

\ref{s:synthetic} below describes the P2D model and its simulation that gives the synthetic data for training the surrogate model. We use the PyBaMM \cite{PyBaMM} library developed at Oxford for the simulations. We also show the driving cycle we simulate and train to, and show the electrochemical effects that must be considered when operating at high current.  We model discharge only, but a similar procedure can be used for charging. In Section~\ref{s:ML} we describe the DL framework we use and details about the training with results in \ref{s:results}. It is shown here that constant discharge data is not sufficient to adequately train the neural network for our high current, variable current driving cycle. The extensions to SOH identification is done in Section~\ref{s:SOH}. We summarize our findings and discuss some promising ideas for future research in a final section. 

\section{P2D Model and Driving Cycle}
\label{s:synthetic}

We train a Deep Neural Network (DNN) on data generated by a P2D model simulation, implemented with the PyBaMM \cite{PyBaMM} library v22.12 with default parameters. A detailed visual and mathematical description of the P2D model can be found in \cite{han2021}. The ensemble of conditions to be modelled involves a battery charged to full capacity, then discharged through a random driving cycle of prescribed currents until it fails to operate at the desired current. The DNN described in the next section is trained on data sampled from the simulation every 100 seconds. The model is trained with predicting the battery voltage and the intercallated Lithium concentrations in electrode particles on a grid in $x$ (scaled position through the electrodes) and $r$ (scaled electrode particle radius). The particle concentrations are internal quantities that are not directly measurable in operation and can only be obtained through the use of electrochemical models. They are necessary to understand and predict the battery behaviour in time dependent high current operation. These model outputs can be combined, due to the deep learning framework we employ, with real operational voltage data to retain the insights from the approximation of these effects from the model as well as correcting for the mismatch between the model and the actual battery. In the current work, we train to the simulation results only.  

The driving cycle consists of choosing random discharge currents every 100 seconds uniformly between zero and 6C. A piecewise linear current profile $I(t)$ is constructed from these values. Batteries are built quite differently for different applications, but generally 1C operation can be modelled well with simple (equivalent circuit) models. It is at higher currents, such as those we consider here, that electrochemical models are needed for accurate voltage predictions. For reference, the maximum constant current that can be drawn over 100s from a fully charged cell modelled by the simulation is 7C (the cell fails in this scenario because of mass transport limitations, not because it has run out of charge). In Figure~\ref{f:f3} below, we show an example of a driving cycle and the corresponding output voltage with the (hidden) quantities of electrolyte ion concentration and potential. To give context to this figure, Lithium intercallated in particles in the left (negative) electrode before the vertical line are leaving the particles, travelling as ions in the electrolyte from left to right, and then intercallating into the right (positive) electrode particles. Between the two vertical lines of the electrolyte is the separator, in which there is only ion transport. On Figure~\ref{f:f3} we also show the results from the SPD model on which our previous, preliminary results \cite{maricela0} were based. 

\begin{figure}
\centerline{
\includegraphics[width=12cm]{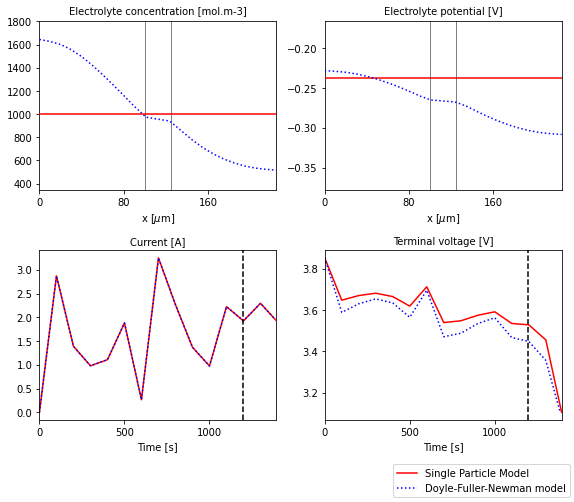}}
\caption{\label{f:f3} An example driving cycle with the current $I(t)$ shown on the bottom left. The battery voltage, an externally measurable quantity, is on the bottom right. The electrolyte concentration and potential are shown on the top lines at the time indicated by the dotted line on the bottom figures. On the top row of figures, the positive electrode is to the left of the first vertical line, the negative electrode is to the right of the second vertical line, and the separator is between the lines.}
\end{figure}

\begin{figure}
\centerline{
\includegraphics[width=8cm]{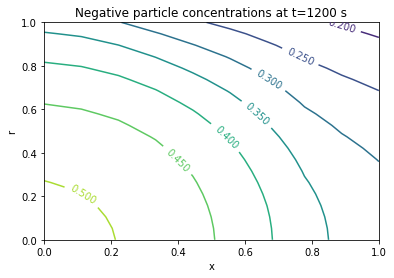} 
\includegraphics[width=8cm]{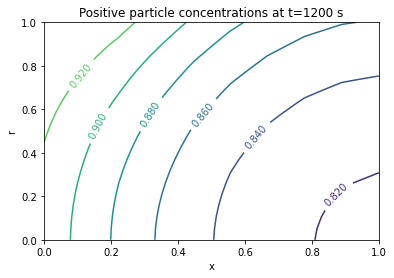} 
}
\caption{\label{f:f57} Intercallated Lithium concentrations in the negative (left) and positive (right) electrodes with scaled electrode width $x$ and particle radius $r$. These are at time 1200 of the driving cycle shown in Figure~\ref{f:f3}, the time shown by a dashed vertical line in that figure.}
\end{figure}

We show the electrode particle concentrations in Figure~\ref{f:f57} at time 1200s of the simulation shown in Figure~\ref{f:f3}. In that previous figure, that time is shown as a dashed vertical line on the graphs on the bottom row. Here we can see the effects captured by the P2D model, the variation in scaled $x$ and $r$ of the intercallated Lithium particles during high current operation. Consider the negative particle concentrations in this Figure. In discharge, intercallated Lithium is leaving these particles. Concentrations near the surface (scaled $r=1$) are lower than at particles centres ($r=0$) since the flux in the particles are limited by diffusion. There is preferred Lithium removal from particles near the separator (scaled $x=1$) since this gives a shorter transport path to the positive electrode. 
These mass transport limiting effects are captured by the P2D model. Since the battery voltage is affected strongly by the surface concentration (scaled $r=1$) these variations are important to capture. In addition, at high currents, battery failure can occur because diffusion of intercallated Lithium cannot keep up with current and scaled surface concentrations are driven to zero (no intercalled Lithium) or one (particles full, cannot accept more Lithium). This can happen while the battery has remaining charge and cannot be captured by simple models. 

The output (particle concentrations and battery voltage) of these simulations is what our neural network model is trained to replace (the model is described in the next section). We imagine that these type of simulations would be run many times by a BMS predicting what could happen in future conditions and warning of failure. The PyBaMM implementation of the P2D (and other) models is excellent, but these models are inherently computationally intensive, unsuitable to real time prediction and control. By replacing them with an DNN surrogate model, for which specialized chips are becoming available \cite{chip1,chip2}, we allow the model to be accessible to a BMS implementation. As mentioned previously, a DL implementation seamlessly allows the addition of experimental data.  

In the next sections, we show how we can replace the P2D dynamics over a time window of 100s with a DNN surrogate model. The DNN takes inputs of initial scaled electrode concentrations on a grid (Figure~\ref{f:f57}), initial voltage, and the linear in time currents from the driving cycle on that time window. Outputs are the final scaled electrode concentrations and voltage. Considering Figure~\ref{f:f57}, the initial particle concentrations depend on the time history of operation and influence future operating behaviour. Note that the time history of the electrolyte concentrations and potentials does not need to be kept as electrolyte dynamics are much faster than particle intercallation \cite{iain}. Thus, the electrolyte potential can be considered to be determinable by the input quantities.  

\section{Deep Learning Approach}
\label{s:ML}
\paragraph{Neural Network Architecture}
Convolutional Neural Networks (CNNs) have enjoyed success on a wide variety of image related tasks. 
 The convolutional layers in a CNN implicitly embed biases into the network architecture that make them especially suited to image based tasks. Convolutional operations respect structural information common to image data, e.g. nearby pixels are related to one another through symmetries and other exploitable relationships. 
 
Profiles of intercalated Lithium concentrations in the electrode particles of a battery can naturally be thought of as two dimensional images, where scaled concentration profiles as a function of the spatial grid and particle radius grid are analogous to pixels. This observation motivates the use of a CNN for the prediction of concentration profiles in this work. 

Concentration profiles for each electrode at time $t$ are the input to a sequence of convolutional layers. The convolved result is concatenated to three additional features, the current requested by the drive cycle at the present time, $I(t)$, in 100 seconds, $I(t+100)$, and the measured voltage $V(t)$. This new set of features is run through a fully connected neural network whose output is a prediction of the voltage and concentration profiles at time $t+100$.

\tikzset{every picture/.style={line width=0.75pt}} 
\begin{figure}[htb!]
\scalebox{.85}{

\begin{tikzpicture}[x=0.75pt,y=0.75pt,yscale=-1,xscale=1]

\draw   (147.28,262.67) -- (139.98,268.88) -- (71.93,253.66) -- (70.16,174.48) -- (77.46,168.27) -- (145.51,183.49) -- cycle ; \draw   (71.93,253.66) -- (79.23,247.44) -- (147.28,262.67) ; \draw   (79.23,247.44) -- (77.46,168.27) ;
\draw   (248.02,281.42) -- (239.26,288.88) -- (157.59,270.61) -- (155.38,171.34) -- (164.14,163.88) -- (245.8,182.15) -- cycle ; \draw   (157.59,270.61) -- (166.36,263.15) -- (248.02,281.42) ; \draw   (166.36,263.15) -- (164.14,163.88) ;
\draw   (360.33,303.66) -- (350.88,311.72) -- (262.75,292) -- (260.24,179.3) -- (269.69,171.25) -- (357.81,190.97) -- cycle ; \draw   (262.75,292) -- (272.21,283.95) -- (360.33,303.66) ; \draw   (272.21,283.95) -- (269.69,171.25) ;
\draw   (62.82,250.07) -- (60.41,252.13) -- (4.28,239.57) -- (2.64,166.14) -- (5.05,164.08) -- (61.18,176.64) -- cycle ; \draw   (4.28,239.57) -- (6.69,237.51) -- (62.82,250.07) ; \draw   (6.69,237.51) -- (5.05,164.08) ;
\draw   (406.3,322.27) -- (403.82,324.39) -- (380.64,319.2) -- (376.69,142.42) -- (379.18,140.3) -- (402.35,145.48) -- cycle ; \draw   (380.64,319.2) -- (383.13,317.09) -- (406.3,322.27) ; \draw   (383.13,317.09) -- (379.18,140.3) ;
\draw   (470.29,381.52) -- (467.8,383.64) -- (444.62,378.46) -- (438.47,103.38) -- (440.96,101.26) -- (464.14,106.45) -- cycle ; \draw   (444.62,378.46) -- (447.1,376.34) -- (470.29,381.52) ; \draw   (447.1,376.34) -- (440.96,101.26) ;
\draw   (512.92,254.22) -- (510.43,256.34) -- (487.24,251.15) -- (482.39,33.74) -- (484.87,31.62) -- (508.06,36.81) -- cycle ; \draw   (487.24,251.15) -- (489.73,249.03) -- (512.92,254.22) ; \draw   (489.73,249.03) -- (484.87,31.62) ;
\draw    (389.58,192.88) -- (440.13,175.44) ;
\draw    (389.58,192.88) -- (441.06,197.86) ;
\draw    (389.58,192.88) -- (441.45,227.89) ;
\draw    (391.45,255.17) -- (442,237.73) ;
\draw    (391.45,255.17) -- (441.45,263.12) ;
\draw    (391.45,255.17) -- (443.06,292.22) ;
\draw    (451.98,129.4) -- (484.79,111.96) ;
\draw    (451.98,129.4) -- (485.4,134.39) ;
\draw    (451.98,129.4) -- (484.79,166.78) ;
\draw    (453.19,191.7) -- (486.01,174.25) ;
\draw    (453.19,191.7) -- (486.61,196.68) ;
\draw    (453.19,191.7) -- (486.01,229.07) ;
\draw   (337.83,313.29) .. controls (336.92,317.86) and (338.75,320.61) .. (343.32,321.52) -- (348.61,322.57) .. controls (355.15,323.87) and (357.96,326.81) .. (357.05,331.39) .. controls (357.96,326.81) and (361.69,325.17) .. (368.23,326.48)(365.29,325.89) -- (373.85,327.59) .. controls (378.42,328.5) and (381.17,326.67) .. (382.08,322.1) ;
\draw   (122.81,270.4) .. controls (122.02,275) and (123.92,277.69) .. (128.52,278.48) -- (134.9,279.58) .. controls (141.47,280.71) and (144.36,283.58) .. (143.57,288.18) .. controls (144.36,283.58) and (148.04,281.85) .. (154.61,282.98)(151.66,282.47) -- (161.39,284.15) .. controls (165.99,284.94) and (168.69,283.04) .. (169.48,278.44) ;
\draw   (229.03,291.08) .. controls (228.24,295.67) and (230.14,298.37) .. (234.73,299.16) -- (241.12,300.26) .. controls (247.69,301.39) and (250.57,304.26) .. (249.78,308.86) .. controls (250.57,304.26) and (254.26,302.53) .. (260.83,303.66)(257.87,303.15) -- (267.61,304.83) .. controls (272.21,305.62) and (274.9,303.72) .. (275.69,299.12) ;
\draw   (276.63,116.5) .. controls (276.83,107.66) and (284.16,100.65) .. (293,100.84) -- (348.81,102) .. controls (357.65,102.19) and (364.65,109.5) .. (364.44,118.34) -- (364.44,118.34) .. controls (364.24,127.17) and (356.91,134.18) .. (348.08,134) -- (292.26,132.83) .. controls (283.43,132.64) and (276.43,125.33) .. (276.63,116.5) -- cycle ;
\draw    (336.85,134.46) -- (375.67,222.29) ;
\draw [shift={(376.47,224.12)}, rotate = 246.16] [color={rgb, 255:red, 0; green, 0; blue, 0 }  ][line width=0.75]    (10.93,-3.29) .. controls (6.95,-1.4) and (3.31,-0.3) .. (0,0) .. controls (3.31,0.3) and (6.95,1.4) .. (10.93,3.29)   ;
\draw [line width=0.75]    (38.54,211.15) -- (67.44,211.93) ;
\draw [shift={(69.44,211.98)}, rotate = 181.54] [color={rgb, 255:red, 0; green, 0; blue, 0 }  ][line width=0.75]    (10.93,-3.29) .. controls (6.95,-1.4) and (3.31,-0.3) .. (0,0) .. controls (3.31,0.3) and (6.95,1.4) .. (10.93,3.29)   ;
\draw [line width=0.75]    (119.98,216.96) -- (152.62,216.96) ;
\draw [shift={(154.62,216.96)}, rotate = 180] [color={rgb, 255:red, 0; green, 0; blue, 0 }  ][line width=0.75]    (10.93,-3.29) .. controls (6.95,-1.4) and (3.31,-0.3) .. (0,0) .. controls (3.31,0.3) and (6.95,1.4) .. (10.93,3.29)   ;
\draw [line width=0.75]    (208.91,218.63) -- (256.53,219.42) ;
\draw [shift={(258.53,219.46)}, rotate = 180.96] [color={rgb, 255:red, 0; green, 0; blue, 0 }  ][line width=0.75]    (10.93,-3.29) .. controls (6.95,-1.4) and (3.31,-0.3) .. (0,0) .. controls (3.31,0.3) and (6.95,1.4) .. (10.93,3.29)   ;
\draw [line width=0.75]    (326.86,223.29) -- (374.47,224.08) ;
\draw [shift={(376.47,224.12)}, rotate = 180.96] [color={rgb, 255:red, 0; green, 0; blue, 0 }  ][line width=0.75]    (10.93,-4.9) .. controls (6.95,-2.3) and (3.31,-0.67) .. (0,0) .. controls (3.31,0.67) and (6.95,2.3) .. (10.93,4.9)   ;
\draw   (516.92,417.22) -- (514.43,419.34) -- (491.24,414.15) -- (488.62,296.57) -- (491.1,294.45) -- (514.29,299.64) -- cycle ; \draw   (491.24,414.15) -- (493.73,412.03) -- (516.92,417.22) ; \draw   (493.73,412.03) -- (491.1,294.45) ;
\draw    (457.19,333.7) -- (490.01,316.25) ;
\draw    (457.19,333.7) -- (490.61,338.68) ;
\draw    (457.19,333.7) -- (490.01,371.07) ;
\draw    (506.19,341.7) -- (537.64,346.38) ;
\draw [shift={(539.61,346.68)}, rotate = 188.48] [color={rgb, 255:red, 0; green, 0; blue, 0 }  ][line width=0.75]    (10.93,-3.29) .. controls (6.95,-1.4) and (3.31,-0.3) .. (0,0) .. controls (3.31,0.3) and (6.95,1.4) .. (10.93,3.29)   ;

\draw (113.4,286.26) node [anchor=north west][inner sep=0.75pt]  [font=\footnotesize,rotate=-12.62] [align=left] {\begin{minipage}[lt]{47.63pt}\setlength\topsep{0pt}
Max Pooling
\begin{center}
3x3
\end{center}

\end{minipage}};
\draw (81.57,172.18) node [anchor=north west][inner sep=0.75pt]  [font=\footnotesize,rotate=-12.62] [align=left] {Convolution};
\draw (178.13,170.65) node [anchor=north west][inner sep=0.75pt]  [font=\footnotesize,rotate=-12.62] [align=left] {Convolution};
\draw (285.96,178.31) node [anchor=north west][inner sep=0.75pt]  [font=\footnotesize,rotate=-12.62] [align=left] {Convolution};
\draw (376.33,122.35) node [anchor=north west][inner sep=0.75pt]  [font=\footnotesize,rotate=-12.62] [align=left] {Dense};
\draw (438.29,84.06) node [anchor=north west][inner sep=0.75pt]  [font=\footnotesize,rotate=-12.62] [align=left] {Dense};
\draw (11.97,172.68) node [anchor=north west][inner sep=0.75pt]  [font=\footnotesize,rotate=-11.46] [align=left] {20x20x2};
\draw (217.62,306.94) node [anchor=north west][inner sep=0.75pt]  [font=\footnotesize,rotate=-12.62] [align=left] {\begin{minipage}[lt]{47.63pt}\setlength\topsep{0pt}
Max Pooling
\begin{center}
3x3
\end{center}

\end{minipage}};
\draw (315,326.62) node [anchor=north west][inner sep=0.75pt]  [font=\footnotesize,rotate=-12.62] [align=left] {\begin{minipage}[lt]{47.63pt}\setlength\topsep{0pt}
Max Pooling
\begin{center}
3x3
\end{center}

\end{minipage}};
\draw (482.55,255.29) node [anchor=north west][inner sep=0.75pt]  [font=\footnotesize,rotate=-11.46] [align=left] {401x2};
\draw (378.05,327.43) node [anchor=north west][inner sep=0.75pt]  [font=\footnotesize,rotate=-11.46] [align=left] {23x2x100};
\draw (435.32,382.77) node [anchor=north west][inner sep=0.75pt]  [font=\footnotesize,rotate=-11.46] [align=left] {100x200};
\draw (284.39,111.69) node [anchor=north west][inner sep=0.75pt]  [font=\footnotesize]  {$I_{t} ,\ \ I_{t+1} ,\ V_{t}$};
\draw (488.55,424.29) node [anchor=north west][inner sep=0.75pt]  [font=\footnotesize,rotate=-11.46] [align=left] {200x1};
\draw (486.29,276.06) node [anchor=north west][inner sep=0.75pt]  [font=\footnotesize,rotate=-12.62] [align=left] {Dense};
\draw (545.29,334.06) node [anchor=north west][inner sep=0.75pt]  [font=\footnotesize,rotate=-12.62] [align=left] {Probabiliby \\of Failure};
\draw (23.57,137.38) node [anchor=north west][inner sep=0.75pt]  [rotate=-12.12]  {$C_{t}^{n,p}$};
\draw (470.57,-0.62) node [anchor=north west][inner sep=0.75pt]  [rotate=-12.12]  {$V_{t+1,\ } C_{t}^{n,p}$};

\end{tikzpicture}}
\caption{Schematic of Network architecture.}\label{fig:CNN}
\end{figure}
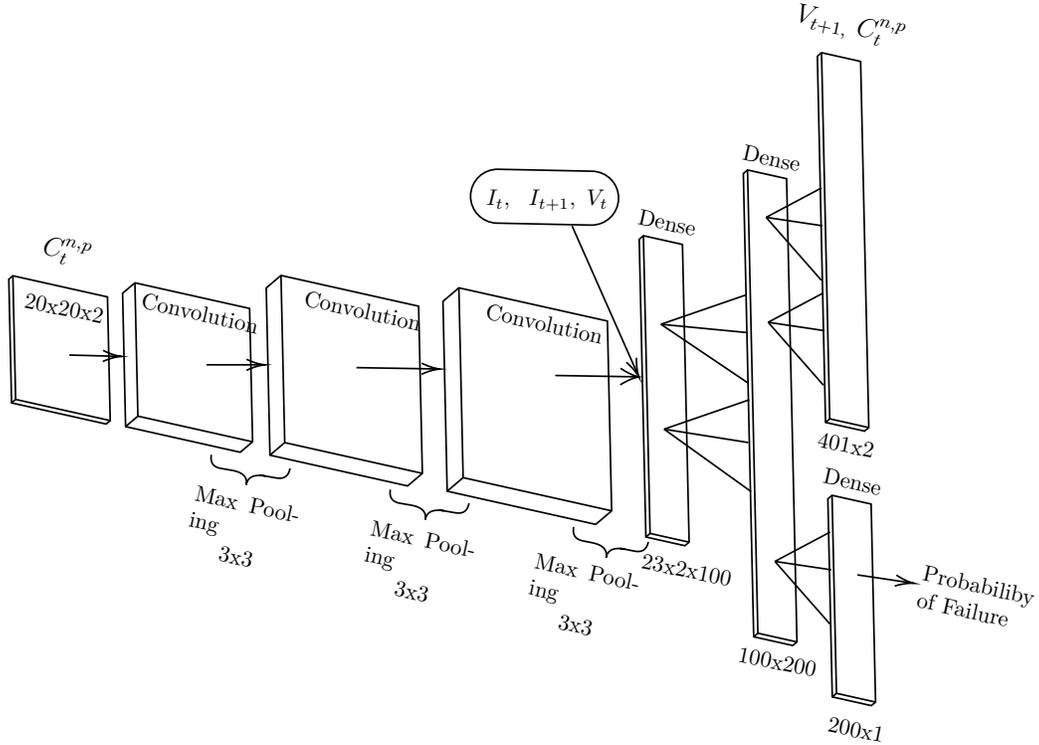

The exact network architecture, depicted in \ref{fig:CNN}, was inspired by AlexNet \cite{Alex}, a popular CNN architecture originally designed to identify images. The network has a total of nine layers. The first six layers consist of three ReLU-activated convolutional layers that are each followed by a Max-Pooling layer. The kernels for the convolutional layers have dimensions 7x7, 5x5, and 3x3. All max-pooling layers have dimension 3x3. Finally, two three-layer fully connected networks are used for prediction; one predicts voltage and flattened concentration profiles, while the other performs a logistic regression to predict the probability that battery operation will fail in the 100-second prediction window. 

\paragraph{Training procedure} 
Sets of training and testing data are generated by running a PyBaMM simulation of the P2D model for a set of 18,000 driving cycles (described in \ref{s:synthetic}). Because the goal is to predict battery concentration profiles, voltages, and the probability of reaching battery discharge 100 seconds into the future, the data set is split into intervals of 100 seconds. A set of 15,000 simulations are used for training, and 3,000 are retained to test the model after training. The model is trained on mini-batches consisting of 64 randomly sampled 100-second intervals. Stochastic optimization is performed using the ADAM optimizer \cite{ADAM}, over 5 epochs, with a decreasing learning rate. Li-ion concentration profiles and voltage accuracy are trained using Mean Squared Error(MSE) as the loss function. Similarly, MSE loss is employed to perform a least squares regression in order to predict the probability of battery discharge in the given 100-second window. Because voltage accuracy is of particular interest, and voltage labels make up one point out of 801 labels, this component of the loss is multiplied by a factor of 10.

\section{Results}\label{s:results}
In this section, we detail the performance of the trained neural network model on the test data set of 3,000 simulations generated via the procedure described in \ref{s:synthetic}. The performance of the model is first investigated in Section~\ref{s:1step} over a 100-second prediction window, in line with how the model is trained. This assumes that concentration profiles and voltages are known at any given time step. In practice, however, concentration profiles are not measurable in a working system. To address this, we report on the predictive accuracy of the model over a K-step prediction horizon
in Section~\ref{s:Kstep}. Starting from a fully charged battery, the model is tasked with predicting the entire discharge trajectory using only its own predicted concentration profiles. The accuracy of the model is assessed for voltage, concentration profile, and discharge failure prediction. 

\subsection{One Step Predictions}
\label{s:1step}

We give some metrics for the DNN model applied to 100 second intervals in the 3,000 test discharge driving cycles in Tables~\ref{table:1} (end interval voltage) and~\ref{t:concentration} (end interval particle concentrations). Inputs are the driving cycle currents $I_{\rm start}$ and $I_{\rm end}$ and the initial voltage and particle concentrations on the $x-r$ grid points. Excellent accuracy is obtained. We show the square root of the mean squared error $l_2$, the mean error $l_1$, and the maximum error $l_\infty$ over all intervals of all 3,000 test discharge cycles. 

\begin{table}[h]
\centering
\begin{tabular}{|l|l|l|l|}
\hline
\textbf{} & \textbf{$l_2$ Error} & \textbf{$l_1$ Error} & \textbf{$l_\infty$ Error} \\ \hline
Mean & $3.73 \times 10^{-5}$ & $3.41 \times 10^{-3}$ & $1.52 \times 10^{-2}$ \\ 
Max  & $6.56 \times 10^{-4}$ & $1.21 \times 10^{-2}$ & $9.39 \times 10^{-2}$ \\ 
\hline
\end{tabular}
\caption{One step voltage errors}
\label{table:1}
\end{table}

\begin{table}[htb!]
\centering
\begin{tabular}{|c|c|c|c|c|}
\hline
\textbf{Concentration} & \textbf{Category} & \textbf{$l_2$ Error} & \textbf{$l_1$ Error} & \textbf{$l_\infty$ Error} \\ \hline
Negative & Mean & $1.52 \times 10^{-4}$ & $8.97 \times 10^{-3}$ & $5.67 \times 10^{-2}$ \\ 
Negative & Max  & $6.14 \times 10^{-4}$ & $1.74 \times 10^{-2}$ & $1.11 \times 10^{-1}$ \\ 
\hline
Positive & Mean & $3.92 \times 10^{-5}$ & $4.37 \times 10^{-3}$ & $2.85 \times 10^{-2}$ \\ 
Positive & Max  & $2.40 \times 10^{-4}$ & $1.08 \times 10^{-2}$ & $6.58 \times 10^{-2}$ \\ 
\hline
\end{tabular}
\caption{One interval concentration errors}
\label{t:concentration}
\end{table}

\subsection{K-step prediction}
\label{s:Kstep}
The results above are for a prediction horizon of 100 seconds. Concentration profiles at a given moment of battery operation are required to make predictions. In this section, we use the model to predict performance over entire discharge cycles. 
Initial concentration profiles will be constant in $x$ and $r$ and can be estimated based on manufacturing specifications. Starting with a fully charged battery, predicted concentrations may be recursively used as inputs to the neural network to predict the entire trajectory of battery operation. Predicted voltages can be employed in the same manner as concentrations. This is a true surrogate for the electrochemical model that can be used for optimization, control, and operation failure prediction.



\begin{figure}[h!]
\centerline{
\includegraphics[width=10cm]{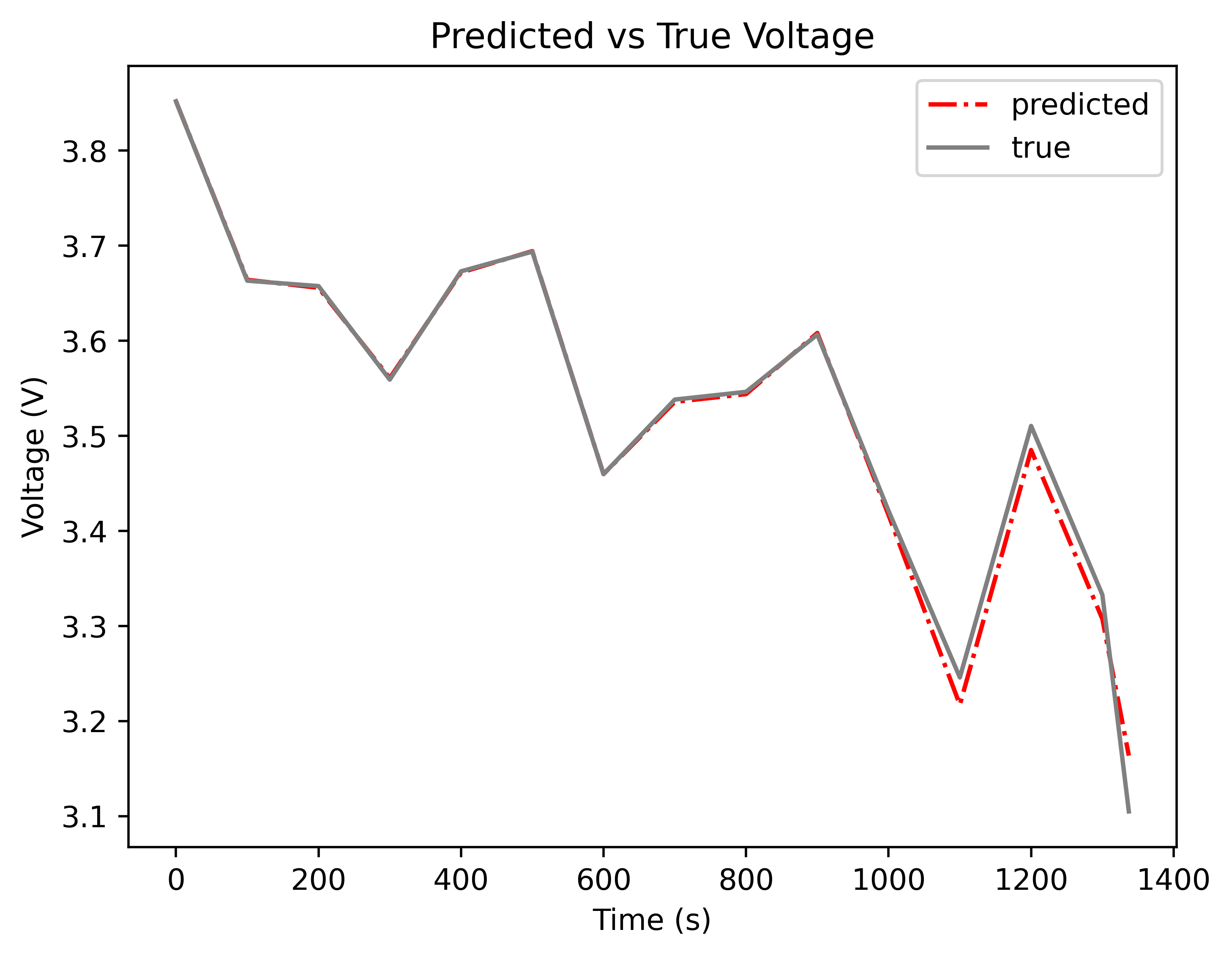}}
\caption{\label{f:meanKstepvolt} The neural network K-step predicted voltage vs the true voltage curve for a simulation with an approximately average $l^\infty$ error over all of the validation data set runs.}
\end{figure}

Error metrics over all intervals of a discharge cycle starting at a fully charged battery are shown in Tables~\ref{table:1K} (voltage) and~\ref{table:2K} (particle concentrations). A representative example of a predicted voltage curve over a driving cycle is shown in Figure~\ref{f:meanKstepvolt}. 

\begin{table}[h]
\centering
\begin{tabular}{|l|l|l|l|}
\hline
\textbf{} & \textbf{$l_2$ Error} & \textbf{$l_1$ Error} & \textbf{$l_\infty$ Error}  \\ \hline
Mean & $5.69 \times 10^{-4}$ & $8.25 \times 10^{-3}$ & $5.98 \times 10^{-2}$ \\ 
Max  & $9.98 \times 10^{-3}$ & $4.19 \times 10^{-2}$ & $4.23 \times 10^{-1}$ \\ 
\hline
\hline
\end{tabular}
\caption{The maximum, minimum, and mean K-step errors in voltage predictions computed over a driving cycle for different metrics.}
\label{table:1K}
\end{table}

\begin{table}[h]
\centering
\begin{tabular}{|c|c|c|c|c|}
\hline
\textbf{Concentration} & \textbf{Category} & \textbf{$l_2$ Error} & \textbf{$l_1$ Error} & \textbf{$l_\infty$ Error} \\ \hline
Negative & Mean & $2.49 \times 10^{-4}$ & $1.16 \times 10^{-2}$ & $5.84 \times 10^{-2}$ \\ 
Negative & Max  & $1.38 \times 10^{-3}$ & $2.74 \times 10^{-2}$ & $1.31 \times 10^{-1}$ \\ 
\hline
Positive & Mean & $1.03 \times 10^{-4}$ & $6.73 \times 10^{-3}$ & $3.52 \times 10^{-2}$ \\ 
Positive & Max  & $7.83 \times 10^{-4}$ & $2.10 \times 10^{-2}$ & $1.07 \times 10^{-1}$ \\ 
\hline
\end{tabular}
\caption{Concentration Errors}
\label{table:2K}
\end{table}

\subsection{Failure Window Prediction}
The neural network model is trained to predict if the battery will reach a voltage cut-off during the 100-second prediction window, meaning it can no longer operate at the desired current. Due to the mass transport limitations at high current operation, this can happen when the battery still has remaining charge. We refer to this state as operation failure. Predictions of battery discharge are compared against labels of 1 for failure and 0 otherwise. A sigmoid activation function is utilized on the network layer that predicts failure to keep network outputs between 0 and 1. This output can be viewed as a prediction of the probability of failure over the next 100 seconds.

In the context of batteries, it is undesirable to reach discharge unexpectedly. This prompts a cautious approach when predicting failure, it is ideal to minimize unpredicted failures (false negatives). Predicting failure early (a false positive), on the other hand, is not ideal but of less consequence. 

In \ref{t:Kstepfailflag}, we report the percentage of false negatives and false positives over the test data set at various likelihood thresholds, e.g. row 1 shows the rate of false negatives and positives when the model estimates that the probability of failure is at least 10\%. 
The model performs well for thresholds at or below 50\%. 
Unsurprisingly, more conservative thresholds lead to higher rates of false positives but fewer false negatives. 
The question of how to balance the risk of false negatives against the wastefulness of false positives is application dependant.  

\begin{table}[hbt!]
  \begin{tabular}{l|l|l}
    Threshold \% & False Negative \% & False Positive \% \\
    \hline
10	    & 0	     &  1.72 \\
20	    & 0.08	 & 1.2 \\
30	    & 0.08	 & 0.8  \\
40	    & 0.16  & 0.72 \\
50      & 0.56   & 0.6
  \end{tabular}
  \caption{K-step prediction of failure interval.}  \label{t:Kstepfailflag}
\end{table}

It is worth briefly noting that due to the nature of the training and testing data sets, battery failures make up a small portion of the training data. For example, consider \ref{f:meanKstepvolt}; this simulation contributes 14 data points to the testing data set, only one of which includes failure.

\subsection{Computational time comparison}

We compared the computational timing of a K-step prediction with the surrogate model vs a Pybamm simulation for the same representative drive cycle. Pybamm takes 23 seconds, while the K-step prediction takes 0.066 seconds. The comparison was performed on  an Intel Xeon CPU with 2 vCPUs (virtual CPUs) and 13GB of RAM.

\subsection{Comparison to constant current data}

\begin{figure}[hbt]
  \includegraphics[width=8cm]{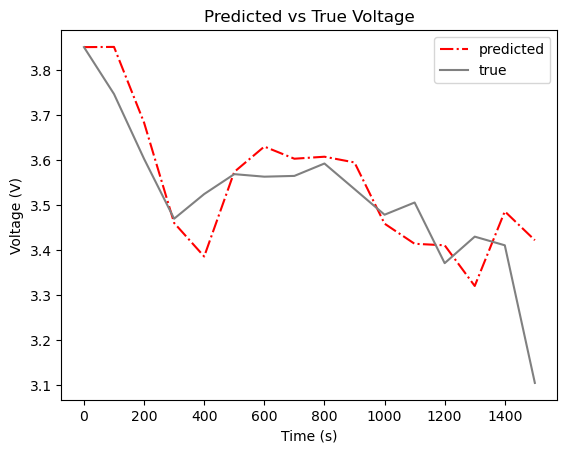}
    \includegraphics[width=8cm]{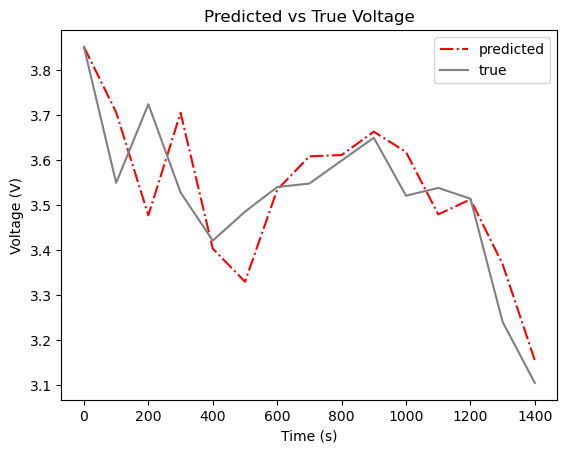}
  \caption{Predictions vs true voltages for a model trained with constant current data}
  \label{f:cc_Verror}
\end{figure}

The DNN model above was trained with data from the driving cycle. We retrain with only constant current discharge data, more typically obtained from discharge experiments. K-step errors in voltage and particle concentrations are shown in Table~\ref{tab:cc_error}. These are more than an order of magnitude larger than the results from the model trained on the variable current driving cycle. Representative voltage errors are shown in Figure~\ref{f:cc_Verror} (compare to Figure~\ref{f:meanKstepvolt}). 

\begin{table}[htb!]
\centering
\begin{tabular}{lccc}
\toprule
\textbf{} & \textbf{$l_2$ Error} & \textbf{$l_1$ Error} & \textbf{$l_\infty$ Error} \\ 
\midrule
\multicolumn{4}{c}{\textbf{Voltage}} \\
\midrule
Mean & \(1.3436 \times 10^{-2}\) & \(8.9579 \times 10^{-2}\) & \(2.5808 \times 10^{-1}\) \\
Max & \(3.9252 \times 10^{-2}\) & \(1.6034 \times 10^{-1}\) & \(4.5668 \times 10^{-1}\) \\
\midrule
\multicolumn{4}{c}{\textbf{Negative Concentration}} \\
\midrule
Mean & \(2.4115 \times 10^{-2}\) & \(1.2968 \times 10^{-1}\) & \(3.8812 \times 10^{-1}\) \\
Max & \(3.7468 \times 10^{-2}\) & \(1.7089 \times 10^{-1}\) & \(4.8327 \times 10^{-1}\) \\
\midrule
\multicolumn{4}{c}{\textbf{Positive Concentration}} \\
\midrule
Mean & \(1.4837 \times 10^{-2}\) & \(1.0049 \times 10^{-1}\) & \(3.2989 \times 10^{-1}\) \\
Max & \(2.2598 \times 10^{-2}\) & \(1.3179 \times 10^{-1}\) & \(4.5263 \times 10^{-1}\) \\
\bottomrule
\end{tabular}
\caption{Errors in Voltage and Concentration prediction for model trained on constant current data}
\label{tab:cc_error}
\end{table}

\section{State of Health Estimation}\label{s:SOH} 

Battery performance degrades with time and use \cite{BMS_SOH1,BMS_SOH2,SOH3,SOH4,SOH5}. 
In this section, we introduce a simple state of health (SOH) parameter $\gamma$ and use the neural network model to estimate this parameter by post-processing the difference between network predictions for a new battery ($\gamma =1$) and simulations of the aged battery ($\gamma < 1$) over a driving cycle. 

Battery SOH degradation can take many forms, including resistance increase, maximum SOC decrease, and changes in the open circuit voltage curve shape. We model the first two of these effects with the parameter $\gamma$. We consider roughly that aging lowers the active particle area of both electrodes to a fraction $\gamma$. At the electrode level, this is equivalent to increasing the current density $I$ for a new battery to $I/\gamma$. We can then consider the response by an aged battery to a driving cycle with currents $I(t)$ to be equivalent to a new battery with currents $I(t)/\gamma$. This also increases electrolyte resistance in a nonlinear way. While these two effects could be scaled independently, this is a simple and convenient SOH model to give a proof of concept for our SOH estimation strategy. 

The neural network model has been trained on batteries with $100\%$ SOH, e.g. $\gamma=1$. Given a battery with a diminished SOH, the goal is to estimate $\gamma$ based on the discrepancy between the predicted K-step voltages and the actual voltages from the aged battery simulation. 
A grid search of values between $[0.75,1.1]$ is performed to estimate the value of $\gamma$. The objective function used for the grid search measures the difference in the voltage curves between the measured and predicted voltages if the predicted failure window matches the actual failure window and is set to $0.5$ otherwise, i.e.

\[
f(\gamma)= \begin{cases}
 \frac{1}{K-1}\sum_{i=1}^{K-1}|V(t_i)_{\text{pred}}-V(t_i)| \quad \text{if failure window correct}\\
 0.5 \quad \text{otherwise}
 \end{cases}.
\]

This process is repeated over five battery cycles. A trimmed mean that cuts $20\%$ of the tails of the distribution of the five estimates is then chosen as the final estimate of the true SOH. Results of this estimation process are shown in \ref{t:SOH estimation}. For each $\gamma < 1$, $\gamma$ is estimated based on discrepancies between predicted voltages and true voltages for 5 randomly generated drive cycles. This corresponds to estimating the SOH of an unknown aged battery after $5$ cycles of operation. To obtain a more comprehensive understanding of the accuracy of the estimation process, estimation trials are conducted five separate times for each gamma value. The predictions have high accuracy. 

\begin{table}[hbt!]
\centering
  \begin{tabular}{l||l|l|l|l|l}
    $\gamma$  & 1 & 2 &3  &4 &5 \\
    \hline
    \hline
$0.80$&$ 0.800      $&$ 0.789$&$ 0.803 $&$ 0.803 $&$ 0.800$\\
$0.85$&$ 0.850      $&$ 0.853 $&$ 0.823 $&$ 0.840       $&$ 0.856$\\
$0.90$&$ 0.906$&$ 0.903 $&$ 0.913 $&$ 0.890       $&$ 0.910$\\
$0.95$&$ 0.953 $&$ 0.973 $&$ 
0.966 $&$ 0.963 $&$ 0.963 $\\
\end{tabular}
  \caption{SOH parameter $\gamma$ vs the estimated value. Estimations are made over 5 battery cycles. The selected cycles are randomly chosen simulations from a pool of 180. This estimation procedure is repeated 5 times.}  \label{t:SOH estimation}
\end{table}

\section{Summary and Future Work}
\label{s:summary}

We have shown that a battery whose operation can be described accurately with a P2D model can be accurately simulated with a computationally inexpensive neural network surrogate model trained on driving cycles comparable to how the battery will be used. The low computational complexity and expense would allow real time predictions and control. The Deep Learning framework can incorporate SOH parameter identification. 

Deep Learning allows integration of data from many sources. Experimental data from real operation can be added to the simulated data used here (with roughly fit model parameters) to give better accuracy. Looked at the other way, simulation data can supplement experimental data, allowing accurate neural network approximation even when the experimental regimes are incomplete. The current work shows the promise of this approach. More realistic SOH effects can be incorporated into the simulations and added as inputs to the DL model and adjusted to match measured voltages. 

\newpage
\bibliographystyle{abbrv}
\bibliography{M1_v2}

\begin{thebibliography}{10}

\bibitem{BMS2}
Y.~Barsukov and J.~Qian.
\newblock {\em Battery Power Management for Portable Devices}.
\newblock Artech House, Boston, 2013.

\bibitem{SOH4}
M.~Berecibar, I.~Gandiaga, I.~Villarreal, N.~Omar, J.~Van~Mierlo, and
  P.~Van~den Bossche.
\newblock {Critical review of state of health estimation methods of Li-ion
  batteries for real applications}.
\newblock {\em Renewable and Sustainable Energy Reviews}, 56(C):572--587, 2016.

\bibitem{BatteryU}
I.~Buchmann.
\newblock Batteries in a portable world: {{A}} handbook on rechargeable
  batteries for non-engineers.
\newblock pages 59--71. {Cadex Electronics Inc.}, 4 edition, 2016.

\bibitem{LIB_overview3}
Z.~P. Cano, D.~Banham, S.~Ye, A.~Hintennach, J.~Lu, M.~Fowler, and Z.~Chen.
\newblock Batteries and fuel cells for emerging electric vehicle markets.
\newblock {\em Nature Energy}, 3(4):279--289, 2018.

\bibitem{P2D}
M.~Doyle, T.~F. Fuller, and J.~Newman.
\newblock Modeling of galvanostatic charge and discharge of the
  {{Lithium}}/{{Polymer}}/{{Insertion}} cell.
\newblock {\em Journal of The Electrochemical Society}, 140(6):1526--1533,
  1993.

\bibitem{ML3}
A.~Geslin, B.~{van Vlijmen}, X.~Cui, A.~Bhargava, P.~A. Asinger, R.~D. Braatz,
  and W.~C. Chueh.
\newblock Selecting the appropriate features in battery lifetime predictions.
\newblock {\em Joule}, 7(9):1956--1965, 2023.

\bibitem{BMS_SOH2}
Z.~Guo, X.~Qiu, G.~Hou, B.~Y. Liaw, and C.~Zhang.
\newblock State of health estimation for lithium ion batteries based on
  charging curves.
\newblock {\em Journal of Power Sources}, 249:457--462, 2014.

\bibitem{han2021}
R.~Han, C.~Macdonald, and B.~Wetton.
\newblock A fast solver for the pseudo-two-dimensional model of lithium-ion
  batteries.
\newblock https://arxiv.org/abs/2111.09251, 2021.

\bibitem{chip1}
B.~Hickmann, J.~Chen, M.~Rotzin, A.~Yang, M.~Urbanski, and S.~Avancha.
\newblock Intel nervana neural network processor-t (nnp-t) fused floating point
  many-term dot product.
\newblock In {\em 2020 IEEE 27th Symposium on Computer Arithmetic (ARITH)},
  pages 133--136, Los Alamitos, CA, USA, jun 2020. IEEE Computer Society.

\bibitem{Equiv1}
h.~Hongwen, R.~Xiong, and F.~Jinxin.
\newblock Evaluation of lithium-ion battery equivalent circuit models for state
  of charge estimation by an experimental approach.
\newblock {\em Energies}, 4:582--598, Dec. 2011.

\bibitem{P2Dreview}
A.~Jokar, B.~Rajabloo, M.~D{\'e}silets, and M.~Lacroix.
\newblock Review of simplified {{Pseudo-two-Dimensional}} models of lithium-ion
  batteries.
\newblock {\em Journal of Power Sources}, 327:44--55, 2016.

\bibitem{chip2}
N.~P. Jouppi and A.~Swing.
\newblock A machine learning supercomputer with an optically reconfigurable
  interconnect and embeddings support.
\newblock In {\em 2023 IEEE Hot Chips 35 Symposium (HCS)}, pages 1--24, Los
  Alamitos, CA, USA, aug 2023. IEEE Computer Society.

\bibitem{ADAM}
D.~P. Kingma and J.~Ba.
\newblock Adam: {{A}} method for stochastic optimization.
\newblock {\em arXiv preprint arXiv:1412.6980}, 2014.

\bibitem{Alex}
A.~Krizhevsky, I.~Sutskever, and G.~E. Hinton.
\newblock Imagenet classification with deep convolutional neural networks.
\newblock {\em Commun. ACM}, 60(6):84–90, may 2017.

\bibitem{LIB_overview2}
M.~Li, J.~Lu, Z.~Chen, and K.~Amine.
\newblock 30 years of lithium-ion batteries.
\newblock {\em Advanced Materials}, 30(33):1800561, 2018.

\bibitem{BMS3}
Q.~Lin, J.~Wang, R.~Xiong, W.~Shen, and H.~He.
\newblock Towards a smarter battery management system: {{A}} critical review on
  optimal charging methods of lithium ion batteries.
\newblock {\em Energy}, 2019.

\bibitem{maricela0}
M.~B. Mckay, B.~Wetton, and R.~B. Gopaluni.
\newblock Learning physics based models of {{Lithium-ion Batteries}}.
\newblock {\em IFAC-PapersOnLine}, 54(3):97--102, 2021.

\bibitem{iain}
I.~R. Moyles, M.~G. Hennessy, T.~G. Myers, and B.~R. Wetton.
\newblock Asymptotic reduction of a porous electrode model for lithium-ion
  batteries.
\newblock {\em SIAM Journal on Applied Mathematics}, 79(4):1528--1549, 2019.

\bibitem{SOH5}
N.~Noura, L.~Boulon, and S.~Jemei.
\newblock A review of battery state of health estimation methods: Hybrid
  electric vehicle challenges.
\newblock {\em World Electric Vehicle Journal}, 2020.

\bibitem{BMS}
S.~K. Pradhan and B.~Chakraborty.
\newblock Battery management strategies: An essential review for battery state
  of health monitoring techniques.
\newblock {\em Journal of Energy Storage}, 51:104427, 2022.

\bibitem{BMS_SOH1}
J.~Remmlinger, M.~Buchholz, M.~Meiler, P.~Bernreuter, and K.~Dietmayer.
\newblock State-of-health monitoring of lithium-ion batteries in electric
  vehicles by on-board internal resistance estimation.
\newblock {\em Journal of Power Sources}, 196(12):5357--5363, 2011.

\bibitem{ML2}
K.~Severson, P.~Attia, N.~Jin, N.~Perkins, B.~Jiang, Z.~Yang, M.~Chen,
  M.~Aykol, P.~Herring, D.~Fraggedakis, M.~Bazant, S.~Harris, W.~Chueh, and
  R.~Braatz.
\newblock Data-driven prediction of battery cycle life before capacity
  degradation.
\newblock {\em Nature Energy}, 4:1--9, 05 2019.

\bibitem{PyBaMM}
V.~Sulzer, S.~G. Marquis, R.~Timms, M.~Robinson, and S.~J. Chapman.
\newblock Python battery mathematical modelling ({{PyBaMM}}).
\newblock {\em Journal of Open Research Software}, 9(1), 2021.

\bibitem{LIB_overview}
J.~M. Tarascon and M.~Armand.
\newblock Issues and challenges facing rechargeable lithium batteries.
\newblock {\em Nature}, 414(6861):359--367, 2001.

\bibitem{bhushan}
M.~Torchio, L.~Magni, B.~Gopaluni, R.~Braatz, and D.~Raimondo.
\newblock {{LIONSIMBA}}: {{A}} matlab framework based on a finite volume model
  suitable for li-ion battery design, simulation, and control.
\newblock {\em Journal of The Electrochemical Society}, 163:A1192--A1205, Apr.
  2016.

\bibitem{ML1}
J.~Zhu, Y.~Huang, M.~Knapp, X.~Liu, Y.~Wang, R.~Gopaluni, Y.~Cao, M.~Heere,
  M.~Mühlbauer, L.~Mereacre, H.~Dai, A.~Senyshyn, X.~Wei, and H.~Ehrenberg.
\newblock Data-driven lithium-ion battery capacity estimation from voltage
  relaxation.
\newblock {\em Nat Commun}, 13:2261, 2022.

\bibitem{SOH3}
Y.~Zou, X.~Hu, H.~Ma, and S.~E. Li.
\newblock Combined state of charge and state of health estimation over
  lithium-ion battery cell cycle lifespan for electric vehicles.
\newblock {\em Journal of Power Sources}, 273:793--803, 2015.

\end{thebibliography}

\end{document}